\documentclass[runningheads]{llncs}
\usepackage{graphicx}
\usepackage{amsmath,amssymb} 
\usepackage{color}
\usepackage{footnote}
\usepackage{array}

\begin{document}
\title{Neural Stereoscopic Image Style Transfer} 

\titlerunning{Neural Stereoscopic Image Style Transfer}
%
\authorrunning{X. Gong et al.}

\author{Xinyu Gong\thanks{Work done while Xinyu Gong was a Research Intern with Tencent AI Lab.}$^\ddagger$
 \quad Haozhi Huang$^\dagger$ \quad  Lin Ma$^\dagger$ \quad  Fumin Shen$^\ddagger$ \\  Wei Liu$^\dagger$ \quad Tong Zhang$^\dagger$ \\
\tt\small{\{neoxygong,huanghz08,forest.linma,fumin.shen\}@gmail.com } \\ 
\tt\small{wl2223@columbia.edu} \\
\tt\small{tongzhang@tongzhang-ml.org}
}
\institute{$^\dagger$Tencent AI Lab \\  $^\ddagger$University of Electronic Science and Technology of China}

\newcommand{\etal}{\textit{et al}.}

\newcolumntype{L}[1]{>{\raggedright\let\newline\\\arraybackslash\hspace{0pt}}m{#1}}
\newcolumntype{C}[1]{>{\centering\let\newline\\\arraybackslash\hspace{0pt}}m{#1}}
\newcolumntype{R}[1]{>{\raggedleft\let\newline\\\arraybackslash\hspace{0pt}}m{#1}}

\maketitle
\renewcommand{\thefootnote}{\fnsymbol{footnote}}

\begin{abstract}
Neural style transfer is an emerging technique which is able to endow daily-life images with attractive artistic styles. Previous work has succeeded in applying convolutional neural networks (CNNs) to style transfer for monocular images or videos. However, style transfer for stereoscopic images is still a missing piece. Different from processing a monocular image, the two views of a stylized stereoscopic pair are required to be consistent to provide observers a comfortable visual experience. In this paper, we propose a novel dual path network for view-consistent style transfer on stereoscopic images. While each view of the stereoscopic pair is processed in an individual path, a novel feature aggregation strategy is proposed to effectively share information between the two paths. Besides a traditional perceptual loss being used for controlling the style transfer quality in each view, a multi-layer view loss is leveraged to enforce the network to coordinate the learning of both the paths to generate view-consistent stylized results. Extensive experiments show that, compared against previous methods, our proposed model can produce stylized stereoscopic images which achieve decent view consistency.
\end{abstract}

\section{Introduction}
With the advancement of technologies, more and more novel devices provide people various visual experiences. Among them, a device providing an immersive visual experience is one of the most popular, including virtual reality devices~\cite{htcvive}, augmented reality devices~\cite{holo}, 3D movie systems~\cite{imax}, and 3D televisions~\cite{3dtv}. A common component shared by these devices is the stereo imaging technique, which creates the illusion of depth in a stereo pair by means of stereopsis for binocular vision. To provide more appealing visual experiences, lots of studies strive to apply engrossing visual effects to stereoscopic images~\cite{Basha2011Geometrically,luo2015geometrically,chang2011content}. Neural style transfer is one of the emerging techniques that can be used to achieve this goal.

\begin{figure}[t]
\begin{center}
 \includegraphics[width=0.7\linewidth]{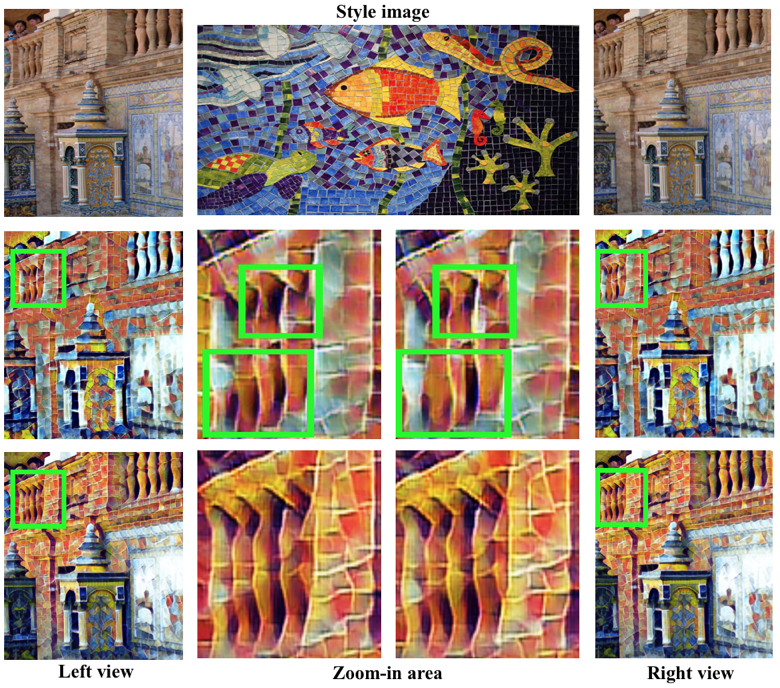}
\end{center}
   \caption{Style transfer applied on stereoscopic images with and without view consistency. The first row shows two input stereoscopic images and one reference style image. The second row includes the stylized results generated by Johnson \etal's method~\cite{johnson2016perceptual}. The middle columns show the zoom-in results, where apparent inconsistency appears in Johnson \etal's method, while our results showed in the third row maintain high consistency.}
\label{fig:show}
\end{figure}

Style transfer is a longstanding problem aiming to combine the content of one image with the style of another. Recently, Gatys \etal~\cite{gatys2016image} revisited this problem and proposed an optimization-based solution utilizing features extracted by a pre-trained convolutional neural network, dubbed Neural Style Transfer, which generates the most fascinating results ever. Following this pioneering work, lots of efforts have been devoted to boosting speed~\cite{johnson2016perceptual,ulyanov2016texture}, improving quality~\cite{ulyanov2016instance,Wang_2017_CVPR}, extending to videos~\cite{gupta2017characterizing,huang2017real,chen2017coherent}, and modeling multiple styles simultaneously~\cite{Huang_2017_ICCV,wang2017zm,li2017diversified}. However, the possibility of applying neural style transfer to stereoscopic images has not yet been sufficiently explored. For stereoscopic images, one straightforward solution is to apply single-image style transfer~\cite{johnson2016perceptual} to the left view and right view separately. However, this method will introduce severe view inconsistency which disturbs the original depth information incorporated in the stereo pair and thus brings observers an uncomfortable visual experience~\cite{kooi2004visual}. Here view inconsistency means that the stylized stereo pair has different stereo mappings from the input. This is because single image style transfer is highly unstable. A slight difference between the input stereo pair may be enormously amplified in the stylized results. An example is shown in the second row of Fig. \ref{fig:show}, where stylized patterns of the same part in the two views are obviously inconsistent.

In the literature of stereoscopic image editing, a number of methods have been proposed to satisfy the need of maintaining view consistency. However, they introduce visible artifacts~\cite{northam2012consistent} and require precise stereo matchings~\cite{Basha2011Geometrically}, while being computationally expensive~\cite{luo2015geometrically}. An intuitive approach is to run single-image style transfer on the left view, and then warp the result according to the estimated disparity to generate the style transfer of the right view. However, this will introduce extremely annoying black regions due to the occluded regions in a stereo pair. Even if filling the black regions with the right-view stylized result, severe edge artifacts are still inevitable.

In this paper, we propose a novel dual path convolutional neural network for the stereoscopic style transfer, which can generate view-consistent high-quality stylized stereo image pairs. Our model takes a pair of stereoscopic images as input simultaneously and stylizes each view of the stereo pair through an individual path.  The intermediate features of one path are aggregated with the features from the other path via a trainable feature aggregation block. Specifically, a gating operation is directly learned by the network to guide the feature aggregation process. Various feature aggregation strategies are explored to demonstrate the superiority of our proposed feature aggregation block. Besides the traditional perceptual loss used in the style transfer for monocular images~\cite{johnson2016perceptual}, a multi-layer view loss is leveraged to constrain the stylized outputs of both views to be consistent in multiple scales. Employing the proposed view loss, our network is able to coordinate the training of both the paths and guide the feature aggregation block to learn the optimal feature fusion strategy for generating view-consistent stylized stereo image pairs. Compared against previous methods, our method can produce view-consistent stylized results, while achieving competitive quality.

In general, the main contributions of our paper are as follows:
\begin{itemize}
\item We propose a novel dual path network for stereoscopic style transfer, which can simultaneously stylize a pair of stereoscopic images while maintaining view consistency. 
\item A multi-layer view loss is proposed to coordinate the training of the two paths of our network, enabling the model, specifically the dual path network, to yield view-consistent stylized results.
\item A feature aggregation block is proposed to learn a proper feature fusion strategy for improving the view consistency of the stylized results.
\end{itemize}

\section{Related Work}
In this work, we try to generate view-consistent stylized stereo pairs via a dual path network, which is closely related to the existing literature on style transfer and stereoscopic image editing.

\smallskip
\noindent\textbf{Neural Style Transfer.}
The first neural style transfer method was proposed by Gatys {\em et~al.}~\cite{gatys2016image}, which iteratively optimizes the input image to minimize a content loss and a style loss defined on a pretrained deep neural network. Although this method achieves fascinating results for arbitrary styles, it is time consuming due to the optimization process. Afterwards, models based on feed-forward CNNs were proposed to boost the speed~\cite{johnson2016perceptual,ulyanov2016texture}, which obtain real-time performance without sacrificing too much style quality. Recently, efforts have been devoted to extending singe-image neural style transfer to videos~\cite{ruder2016artistic,Huang_2017_ICCV,chen2017coherent}. The main challenge for video neural style transfer lies in preventing flicker artifacts brought by temporal inconsistency. To solve this problem, Ruder \etal~\cite{ruder2016artistic} introduced a temporal loss to the time-consuming optimization-based method proposed by Gatys \etal~\cite{gatys2016image}. By incorporating temporal consistency into a feed-forward CNN in the training phase, Huang \etal~\cite{huang2017real} were able to generate temporally coherent stylized videos in real time. Gupta {\em et~al.}~\cite{gupta2017characterizing} also accomplished real-time video neural style transfer by a recurrent convolutional network trained with a temporal loss. Besides the extensive literature on neural style transfer for images or videos, there is still a short of studies on stereoscopic style transfer. Applying single-image style transfer on stereoscopic images directly will cause view inconsistency, which provides observers an uncomfortable visual experience. In this paper, we propose a dual path network to share information between both views, which can accomplish view-consistent stereoscopic style transfer.

\smallskip
\noindent\textbf{Stereoscopic Image Editing.}
The main difficulty of stereoscopic image editing lies in maintaining the view consistency. Basha {\em et~al.}~\cite{Basha2011Geometrically} successfully extended single image seam carving to stereoscopic images, by considering visibility relationships between pixels. A patch-based synthesis framework was presented by Luo {\em et~al.}~\cite{luo2015geometrically} for stereoscopic images, which suggests a joint patch-pair search to enhance the view consistency. Lee {\em et~al.}~\cite{lee2012scene} proposed a layer-based stereoscopic image resizing method, leveraging image warping to handle the view correlation. In~\cite{northam2012consistent}, Northam \etal \, proposed a view-consistent stylization method for simple image filters, but introducing severe artifacts due to layer-wise operations. Kim \etal~\cite{kim2013stereoscopic} presented a projection based stylization method for stereoscopic 3D lines, which maps stroke textures information through the linked parameterized stroke paths in each view. Stavrakis \etal~\cite{stavrakis2005image} proposed a warping based image stylization method, warping the left view of the stylized image to the right and using a segment merging operation to fill the occluded regions. The above methods are either task specific or time-consuming, which are not able to generalize to the neural style transfer problem. In this paper, we incorporate view consistency into the training phase of a dual path convolutional neural network, thus generating view-consistent style transfer results with very high efficiency. 

\section{Proposed Method}
\begin{figure}[t]
\begin{center}
   \includegraphics[width=0.6\linewidth]{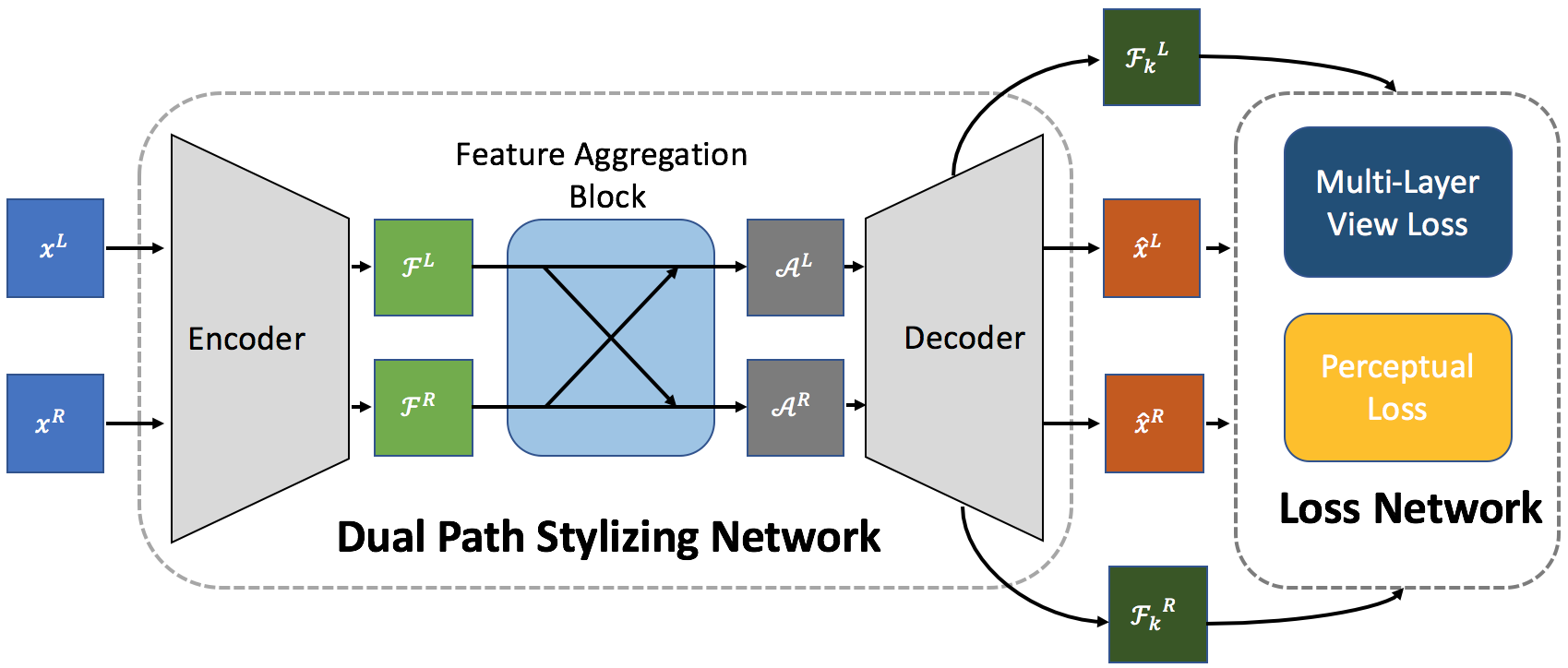}
\end{center}
   \caption{An overview of our proposed model, which consists of a dual path stylizing network and a loss network. The dual path stylizing network takes a pair of stereoscopic images $x^L$ and $x^R$ as input, generating the corresponding stylized images $\widehat{x}^L$ and $\widehat{x}^R$. A feature aggregation block is proposed to share information between the two paths. The loss network calculates the perceptual loss and the multi-layer view loss to guide the training of the stylizing network.}
\label{fig:o}
\end{figure}

Generally, our model is composed of two parts: a dual path stylizing network and a loss network (see Fig. \ref{fig:o}). The dual path stylizing network takes a stereo pair and processes each view in an individual path. A feature aggregation block is embedded into the stylizing network to effectively share feature level information between the two paths. The loss network computes a perceptual loss and a multi-layer view loss to coordinate the training of both the paths of the stylizing network for generating view-consistent stylized results.

\subsection{Dual Path Stylizing Network}
\begin{figure}[t]
\begin{center}
   \includegraphics[width=0.7\linewidth]{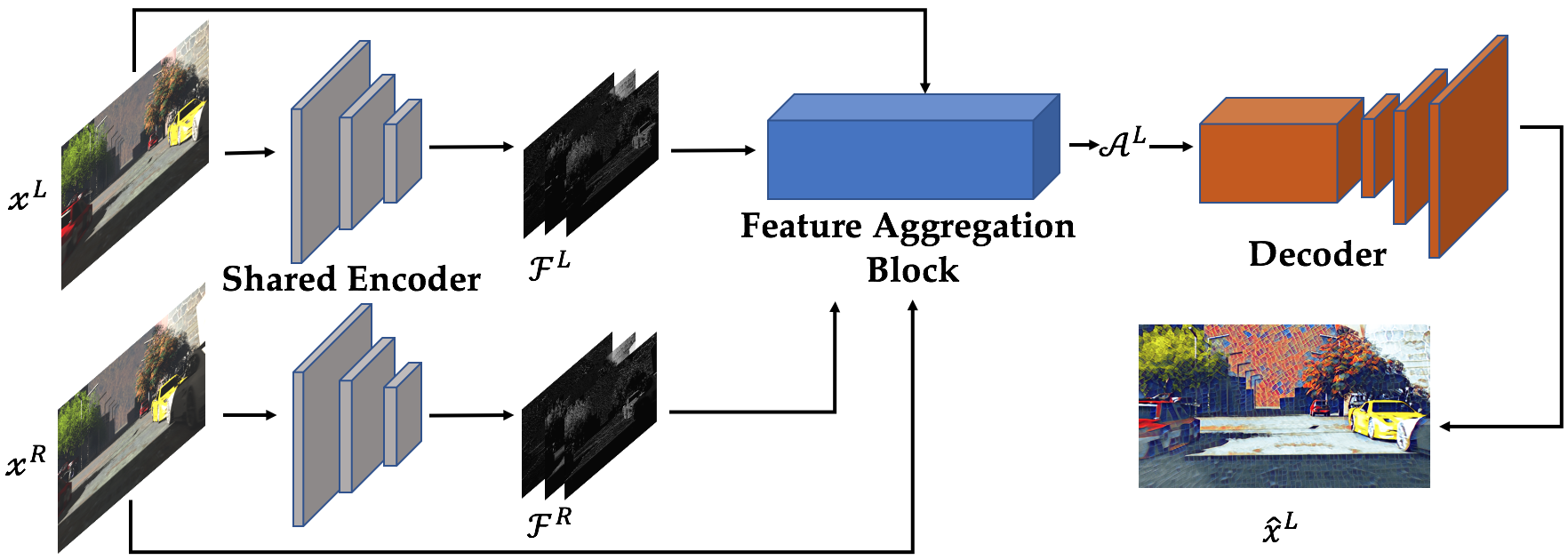}
\end{center}
   \caption{The architecture of the stylizing network, consisting of an encoder, a feature aggregation block, and a decoder. Input images $x^L$ and $x^R$ are encoded to yield the feature maps $\mathcal{F}^L$ and $\mathcal{F}^R$. The feature aggregation block takes $\mathcal{F}^L$ and $\mathcal{F}^R$ as input and aggregates them into $\mathcal{A}^L$. Then $\mathcal{A}^L$ is decoded to yield the stylized result $\widehat{x}^L$.}
\label{fig:sty}
\end{figure}

Our stylizing network is composed of three parts: an encoder, a feature aggregation block, and a decoder. 
The architecture of the stylizing network is shown in Fig. \ref{fig:sty}.  For simplicity, we mainly illustrate the stylizing process of the left view, which is identical to that of the right view. 
First, the encoder, which is shared by both paths, takes the original images as input and extracts initial feature maps $\mathcal{F}^L$ and $\mathcal{F}^R$ for both views. Second, in the feature aggregation block,  $\mathcal{F}^L$ and $\mathcal{F}^R$ are combined together to formulate an aggregated feature map $\mathcal{A}^L$. Finally, $\mathcal{A}^L$ is decoded to produce the stylized image of the left view $\widehat{x}^L$.

\subsubsection{Encoder-decoder.} 
Our encoder downsamples the input images, and extracts the corresponding features progressively. The extracted features are then fed to the feature aggregation block. Finally, our decoder takes the aggregated feature map $\mathcal{A}^L$ as input, and decodes it into stylized images. Note that the encoder and decoder are shared by both views. The specific architectures of the encoder and decoder are shown in Sec. \ref{sec:config}.

\subsubsection{Feature Aggregation Block.}

As aforementioned, separately applying a single-image style transfer algorithm on each view of a stereo image pair will cause view inconsistency. Thus, we introduce a feature aggregation block to integrate the features of both the paths, enabling our model to exploit more information from both views to preserve view consistency.

\begin{figure}[t]
\begin{center}
   \includegraphics[width=0.7\linewidth]{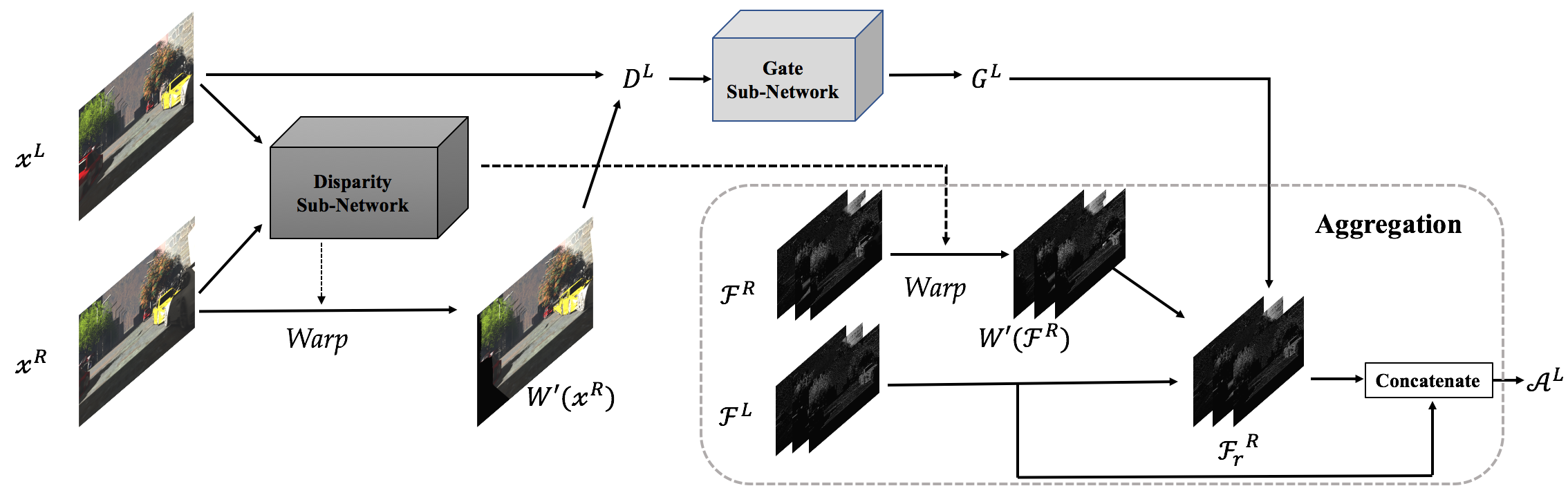}
\end{center}
   \caption{The architecture of the feature aggregation block. The feature aggregation block takes the input stereo pair $x^L$ and $x^R$ and the corresponding encoder's outputs $\mathcal{F}^L$ and $\mathcal{F}^R$. Then, it computes the aggregated feature map $\mathcal{A}^L$. The proposed feature aggregation block consists of three key components: a disparity sub-network, a gate sub-network, and an aggregation.}
\label{fig:agg}
\end{figure}

The architecture of the feature aggregation block is shown in Fig. \ref{fig:agg}. Taking the original stereoscopic images and the features extracted by the encoder as input, the feature aggregation block outputs an aggregated feature map $\mathcal{A}^L$, which absorbs information from both views.

Specifically, a disparity map is predicted by a pretrained disparity sub-network. The predicted disparity map is used to warp the initial right-view feature map $\mathcal{F}^R$ to align with the initial left-view feature map $\mathcal{F}^L$, obtaining the warped right-view feature map $W'(\mathcal{F}^R)$. Explicitly learning a warp operation in this way can reduce the complexity of extracting pixel correspondence information for the model. However, instead of directly concatenating the warped right-view feature map $W'(\mathcal{F}^R)$ with the initial left-view feature map $\mathcal{F}^L$, a gate sub-network is adopted to learn a gating operation for guiding the refinement of $W'(\mathcal{F}^R)$, to generate the refined right feature map $\mathcal{F}_r^R$.  
Finally, we concatenate $\mathcal{F}_r^R$ with $\mathcal{F}^L$ along the channel axis to obtain the aggregated feature map $\mathcal{A}^L$.

\smallskip
\noindent\textbf{Disparity Sub-network.}
Our disparity sub-network takes the concatenation of both views of the stereoscopic pair as input, and outputs the estimated disparity map. It is pretrained on the \emph{Driving} dataset~\cite{MIFDB16} in a supervised way, which contains ground-truth disparity maps.
To predict the disparity map for the left view, both views of the stereoscopic pair are concatenated along the channel axis to formulate $\{x^R, x^L\}$, which is thereafter fed to the disparity sub-network. Similarly, $\{x^L, x^R\}$ is the input for predicting the right disparity map. The specific architecture of our disparity sub-network is shown in Sec. \ref{sec:config}. The architecture of our disparity sub-network is simple; however, it is efficient and does benefit the decrease of the view loss. It is undoubted that applying a more advanced disparity estimation network can boost the performance further at the cost of efficiency, which is out of the scope of this paper.

\smallskip
\noindent\textbf{Gate Sub-network.}
The gate sub-network is proposed to generate a gate map for guiding the refinement of $W'( \mathcal{F}^R)$. First, using bilinear interpolation, we resize the input stereoscopic pair $x^L$, $x^R$ to the same resolution as the initial left-view feature map $\mathcal{F}^L$, which is denoted as $r(x^L)$ and $r(x^R)$. 
Then we calculate the absolute difference between $r(x^L)$ and $W'(r(x^R))$:
\begin{equation}
D^L = \left|r(x^L) - W'(r(x^R))\right|.
\end{equation}
Taking $D^L$ as input, the gate sub-network predicts a single channel gate map $G^L$, which
has the same resolution as $\mathcal{F}^L$.  The range of the pixel values lies in $[0,1]$, which will be used to refine the warped right-view feature map $W'(\mathcal{F}^R)$ later. The specific architecture of the gate sub-network is shown in Sec. \ref{sec:config}.

\smallskip
\noindent\textbf{Aggregation.}
Under the guidance of the gate map generated by the gate sub-network, we refine the warped right-view feature map $W'(\mathcal{F}^R)$ with the initial left-view feature map $\mathcal{F}^L$ to generate a refined right-view feature map:
\begin{equation}
\mathcal{F}_r^R = W'(\mathcal{F}^R)  \odot G^L + \mathcal{F}^L  \odot (1-G^L),
\label{eq:refine}
\end{equation}
where $\odot$ denotes element-wise multiplication. In our experiments, we find that concatenating $W'(\mathcal{F}^R)$ with $\mathcal{F}^L$ directly to formulate the final aggregated left-view feature map $\mathcal{A}^L$ will cause ghost artifacts in the stylized results. This is because the mismatching between $\mathcal{F}^L$ and $W'(\mathcal{F}^R)$ , which is caused by occlusion and inaccurate disparity prediction, will incorrectly introduce right-view information to the left view. Using the gating operation can avoid this issue. Finally, the refined right-view feature map $\mathcal{F}_r^R$ is concatenated with the initial left-view feature map $\mathcal{F}^L$ to formulate the aggregated left-view feature map $\mathcal{A}^L$.

\subsection{Loss Network}
\begin{figure}[t]
\begin{center}
   \includegraphics[width=0.6\linewidth]{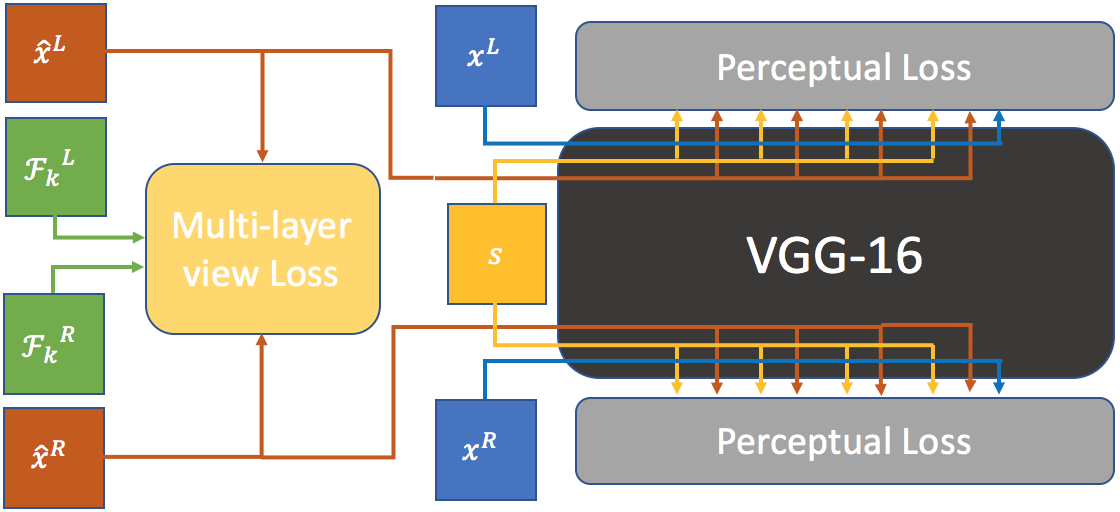}
\end{center}
   \caption{The architecture of the loss network. The perceptual losses of the two views are calculated separately, while the multi-layer view loss is calculated based on the outputs and the features of both views.}
\label{fig:loss}
\end{figure}

Different from the single-image style transfer~\cite{johnson2016perceptual}, the loss network used by our method serves for two purposes. One is to evaluate the style quality of the outputs, and the other is to enforce our network to incorporate view consistency in the training phase. Thus, our loss network calculates a perceptual loss and a multi-layer view loss to guide the training of the stylizing network:
\begin{equation}
\mathcal{L}_{\text{total}} = \sum\limits_{d \in \{ L, R \} }\mathcal{L}_{\text{perceptual}}(s, x^{d}, \widehat{x}^{d}) + \lambda\mathcal{L}_{\text{view}}( \widehat{x}^{L}, \widehat{x}^{R}, \mathcal{F}_k^L, \mathcal{F}_k^R),
\end{equation}
where $\mathcal{F}_k$ denotes the $k$-th layer feature map of the decoder in the stylizing network. $s$ is the reference style image. The architecture of our loss network is shown in Fig. \ref{fig:loss}. While the perceptual losses of the two views are calculated separately, the multi-layer view loss is calculated based on the outputs and the features of both views. By training with the proposed losses, the stylizing network learns to coordinate the training of both the paths to leverage the information from both views, eventually generating stylized and view-consistent results.

\subsubsection{Perceptual Loss.}
We adopt the definition of the perceptual loss in ~\cite{johnson2016perceptual}, which has been demonstrated effective in neural style transfer. The perceptual loss is employed to evaluate the stylizing quality of the outputs, which consists of a content loss and a style loss:
\begin{equation}
\mathcal{L}_{\text{perceptual}}(s, x^{d}, \widehat{x}^{d}) = \alpha\mathcal{L}_{\text{content}}(x^{d}, \widehat{x}^{d}) +\beta\mathcal{L}_{\text{style}}(s, \widehat{x}^{d}),
\end{equation}
where $\alpha$, $\beta$ are the trade-off weights. We adopt a pretrained VGG-16 network~\cite{simonyan2014very} to extract features for calculating the perceptual loss. 

The content loss is introduced to preserve the high-level content information of the inputs:
\begin{equation}
\small
\mathcal{L}_{\text{content}}(x^{d}, \widehat{x}^{d}) = \sum\limits_{l}\frac{1}{H^l  W^l  C^l}\left\Vert\mathcal{F}^{l}(x^{d})- \mathcal{F}^{l}(\widehat{x}^{d})\right\Vert_{2}^{2},
\end{equation}
where $\mathcal{F}^{l}$ denotes the feature map at layer $l$ in the VGG-16 network. $W^l, H^l, C^l$ are the height, width, and channel size of the feature map at layer $l$, respectively. The content loss constrains the feature maps of $x^d$ and $\widehat{x}^{d}$ to be similar, where $d = \{ L,R \}$ represents different views. 

The style loss is employed to evaluate the stylizing quality of the generated images.
Here we use the Gram matrix as the style representation, which has been demonstrated effective in~\cite{gatys2016image}:
\begin{equation}
G^{l}_{ij}(x^d) = \frac{1}{H^l W^l}\sum\limits_{h}^{H^l}\sum\limits_{w}^{W^l}\mathcal{F}^{l}(x^{d})_{h,w,i}\mathcal{F}^{l}(x^{d})_{h,w,j},
\end{equation}
where $G^{l}_{ij}$ denotes the $i,j$-th element of the Gram matrix of the feature map at layer $l$. The style loss is defined as the mean square error between the Gram matrices of the output and the reference style image:
\begin{equation}
\mathcal{L}_{\text{style}}(s, \widehat{x}^{d}) = \sum\limits_{l}\frac{1}{C^l}^2\left\Vert G^l(s)-G^l(\widehat{x}^{d})\right\Vert_{2}^{2}.
\end{equation}
Matching the Gram matrices of feature maps has also been demonstrated to be equivalent to minimizing the Maximum Mean Discrepancy (MMD) between the output and the style reference~\cite{li2017demystifying}.

\subsubsection{Multi-layer View Loss.}
\label{sec:VL}
Besides a perceptual loss, a novel multi-layer view loss is proposed to encode view consistency into our model in the training phase. The definition of the multi-layer view loss is: 
\begin{equation}
\mathcal{L}_{\text{view}} = \mathcal{L}_{\text{view}}^{\text{img}} + \mathcal{L}_{\text{view}}^{\text{feat}},
\end{equation}
where the image-level view loss constrains the outputs to be view-consistent, and the feature-level view loss constrains the feature maps in the stylizing network to be consistent.
The image-level view loss is defined as:
\begin{equation}
\begin{split}
\mathcal{L}_{\text{view}}^{\text{img}} &= \frac{1}{\sum_{i,j}M^L_{i,j}}\left\Vert M^L \odot (\widehat{x}^{L} - W(\widehat{x}^R)) \right\Vert^{2}_2 \\&+  \frac{1}{\sum_{i,j}M^R_{i,j}}\left\Vert M^R \odot (\widehat{x}^{R} - W(\widehat{x}^L)) \right\Vert^{2}_2,
\end{split}
\label{eq:img-view}
\end{equation}
where $M$ is the per-pixel confidence mask of the disparity map, which has the same shape as stylized images. The value of $M_{i,j}$ is either 0 or 1, where 0 in mismatched areas, and 1 in well-matched corresponding areas. $\widehat{x}^L$ and $\widehat{x}^R$ are stylized results. We use $W$ to denote the warp operation using the ground-truth disparity map, provided by the Scene Flow Datasets~\cite{MIFDB16}. Thus, $W(\widehat{x}^L)$ and $W(\widehat{x}^R)$ are a warped stylized stereo pair, using the ground-truth disparity map.

In order to enhance view consistency of stylized images further, we also enforce the corresponding activation values on intermediate feature maps of left and right content images to be identical. Thus, the feature-level view loss is introduced. Similarly, the feature-level view loss is defined as follow:
\begin{equation}
\begin{split}
\mathcal{L}_{\text{view}}^{\text{feat}} &= \frac{1}{\sum_{i,j}m^L_{i,j}}\left\Vert m^L\odot [\mathcal{F}_{k}^{L} - W(\mathcal{F}_{k}^{R})] \right\Vert^{2}_2 \\&+  \frac{1}{\sum_{i,j}m^R_{i,j}}\left\Vert m^R\odot[\mathcal{F}_{k}^{R} - W(\mathcal{F}_{k}^{L})] \right\Vert^{2}_2,
\end{split}
\end{equation}
where $m$ is the resized version of $M$, sharing the same resolution as the $k$-th layer's feature map in the decoder. $\mathcal{F}_{k}^{L}$ and $\mathcal{F}_{k}^{R}$ are the feature maps fetched out from the $k$-th layer in the stylizing network. Similarly, $W(\mathcal{F}_{k}^{L})$ and $W(\mathcal{F}_{k}^{R})$ are the warped feature maps using the ground-truth disparity map.
\section{Experiments}
\subsection{Implementation}
\label{sec:config}
The specific configuration of the encoder and the decoder of our model is shown in Tab. \ref{tab:config}. We use \emph{Conv} to denote Convolution-BatchNorm-Activation  block. $C_{in}$ and $C_{out}$ denote the channel numbers of the input and the output respectively.  \emph{Res} denotes the Residual block, following a similar configuration to~\cite{johnson2016perceptual}. \emph{Deconv} denotes Deconvolution-BatchNorm-Activation block.

\begin{table}
\caption{Model configuration.}
\small
\centering
\resizebox{.7\linewidth}{!}{
\begin{tabular}{ |c|c|c|c|c|c|c| }
\hline
Layer & Kernel & Stride & $C_{in}$ & $C_{out}$ & Acitivation\\ \hline
\multicolumn{6}{ |c| }{Encoder} \\ \hline
Conv & 3$ \times $3 & 1 & 3 & 16 & ReLU \\
Conv & 3$ \times $3 & 2 & 16 & 32 & ReLU \\
Conv & 3$ \times $3 & 2 & 32 & 48 & ReLU \\ 
  &  &  &  &  &  \\ 
  &  &  &  &  &  \\ 
  &  &  &  &  &  \\ 
  &  &  &  &  &  \\
\hline
\multicolumn{6}{ |c| }{Decoder} \\ \hline
Conv & 3$ \times $3 & 1 & 96 & 96 & ReLU \\ 
Conv & 3$ \times $3 & 1 & 96 & 48 & ReLU\\
Res $\times$ 5 &  & & 48 & 48 & ReLU\\
Deconv & 3$ \times $3 & 0.5 & 48 & 32 & ReLU\\
Deconv & 3$ \times $3 & 0.5 & 32 & 16 & ReLU\\
Conv & 3$ \times $3 & 1 & 16 & 3 & tanh\\
\hline
\end{tabular}

\quad

\begin{tabular}{ |c|c|c|c|c|c|c| }
\hline
Layer & Kernel & Stride & $C_{in}$ & $C_{out}$ & Acitivation\\ \hline
\multicolumn{6}{ |c| }{Disparity Sub-network} \\ \hline
Conv & 3$ \times $3 & 1 & 6 & 32 & ReLU \\
Conv & 3$ \times $3 & 2 & 32 & 64 & ReLU \\
Conv & 3$ \times $3 & 2 & 64 & 48 & ReLU \\
Res $\times$ 5 &  & & 48 & 48 & ReLU\\
Deconv & 3$ \times $3 & 0.5 & 48 & 24 & ReLU\\
Deconv & 3$ \times $3 & 0.5 & 24 & 8 & ReLU\\
Conv & 3$ \times $3 & 1 & 8 & 3 & ReLU\\
Conv & 3$ \times $3 & 1 & 3 & 1 & -\\ \hline
\multicolumn{6}{ |c| }{Gate Sub-network} \\ \hline
Conv & 3$ \times $3 & 1 & 3 & 6 & ReLU \\
Conv & 1$ \times $1 & 1 & 6& 12 & ReLU\\
Conv & 1$ \times $1 & 1 & 12 & 6 & ReLU\\
Conv & 1$ \times $1 & 1 & 6 & 3 & ReLU\\
Conv & 1$ \times $1 & 1 & 3 & 1 & tanh\\
\hline
\end{tabular}
}
\label{tab:config}
\end{table}

We use \emph{Driving} in the Scene Flow Datasets~\cite{MIFDB16} as our dataset, which contains $4.4k$ pairs of stereoscopic images. 440 pairs of them are used as testing samples, while the rest are used as training samples. Besides, we also use the stereo images from Flickr~\cite{flickr}, \emph{Driving} test set and Sintel~\cite{Butler:ECCV:2012} to show the visual quality of our results in Sec.\ref{QR}. In addition, images from Waterloo-IVC 3D database~\cite{wang2015quality} are used to conduct our user study. Testing on various datasets in this way demonstrates the  generalization ability of our model. The loss network (VGG-16) is pretrained on the image classification task~\cite{simonyan2014very}. Note that during the training phase, the multi-layer view loss is calculated using the ground-truth disparity map provided by the Scene Flow Datasets~\cite{MIFDB16} to warp fetched feature maps and stylized images. Specifically, we fetch feature maps at $7$-th layer of decoder to calculate feature-level view loss according to our experiments.

The disparity sub-network is first pretrained and fixed thereafter. Then, we train the other parts of the stylizing network for 2 epochs. The input image resolution is $960 \times 540$. We set $\alpha=1$, $\beta=500$, $\lambda = 100$.  The batch size is set to $1$. The learning rate is fixed as $1e-3$. For optimization we use Adam~\cite{kingma2014adam}.

\subsection{Qualitative Results}
\label{QR}
\begin{figure}[t]
\begin{center}
   \includegraphics[width=0.65\linewidth]{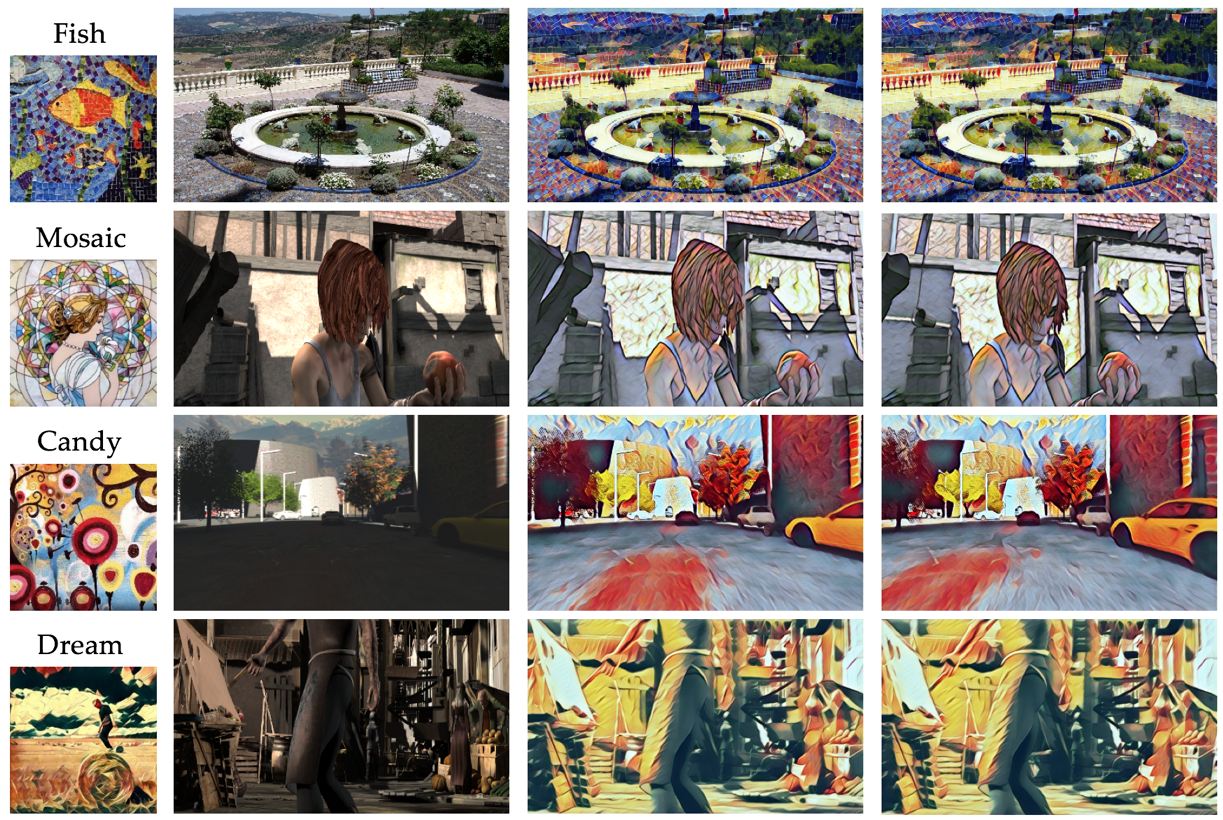}
\end{center}
   \caption{Visual results of our proposed stereoscopic style transfer method. While the high-level contents of the inputs are well preserved, the style details are successfully transferred from the given style images. Meanwhile, view consistency is maintained. }
\label{fig:style}
\end{figure}

We apply the trained model to some stereoscopic pictures from Flickr~\cite{flickr} to show the visual qualities of different styles. In Fig. \ref{fig:style}, stylized results in four different styles are presented, from which we can see that the semantic content of the input images are preserved, while the texture and color are transferred from the reference style images successfully. Besides, view consistency is also maintained.

\subsection{Comparison}
In this section, we compare our method with the single image style transfer method~\cite{johnson2016perceptual}. Though there are many alternative baseline designed for single image neural style transfer, both of them will suffer from similar view inconsistency artifacts as Johnson’s method~\cite{johnson2016perceptual}. Hence, we only choose \cite{johnson2016perceptual} as a representative. Also, we testify the effectiveness of the multi-layer view loss and the feature aggregation block.

As the evaluation metric, we define a term called the mean view loss $MVL$:
\begin{equation}
\begin{split}
MVL = \frac{1}{N}\sum\limits_{n=1}^N \mathcal{L}_{\text{view}}^{\text{img}}(I_n),
\end{split}
\end{equation}
where $N$ is the total number of test images, $I_n$ is the $n$-th image in the test dataset, $\mathcal{L}_{\text{view}}^{\text{img}}$ is the image-level view loss defined in Equation~\ref{eq:img-view}.
In other words, $MVL$ is employed to evaluate the average of the image-level view losses over the whole test dataset.
Similarly, we also define mean style loss ($MSL$) and mean content loss ($MCL$) :
\begin{equation}
MSL = \frac{1}{N}\sum\limits_{n=1}^N \mathcal{L}_{\text{style}}(I_n),
\end{equation}
\begin{equation}
MCL = \frac{1}{N}\sum\limits_{n=1}^N \mathcal{L}_{\text{content}}(I_n).
\end{equation}

For clarity, the single image style transfer method is named as \emph{SingleImage}, where the single image method trained with image-level view loss is named as \emph{SingleImage-IV}. 
Our full model with a feature aggregation block trained with a multi-layer view loss is named as \emph{Stereo-FA-MV}.
The variant model with a feature aggregation block but trained with an image-level view loss is named as \emph{Stereo-FA-IV}.
We evaluate the $MVL$, $MSL$ and $MCL$ of the above models across four styles: \emph{Fish}, \emph{Mosaic}, \emph{Candy} and \emph{Dream}, where the MSLs are coordinated into a similar level.
In Tab. \ref{tab:eval1}, we can see that the mean view loss $MVL$ of our full model \emph{Stereo-FA-MV} is the smallest. The result of the single image style transfer method is the worst. 
Comparing \emph{Stereo-FA-IV} with \emph{SingleImage-IV}, we know that the feature aggregation block benefits the view consistency. Comparing \emph{Stereo-FA-MV} with \emph{Stereo-FA-IV}, we find that constraining the view loss in the feature level besides the image level improves the view consistency further. We also conduct the experiment with fine-tuning the whole network together instead of freezing the disparity sub-network \emph{Stereo-FA-dp-IV}, which performs comparably with \emph{Stereo-FA-IV}.

\begin{table}[t]
\caption{ $MVL$, $MSL$ and $MCL$ of five different models over 4 styles, where $MSL$s are coordinated into a similar level.}
\centering
\resizebox{.9\linewidth}{!}{
    \begin{tabular}{ C{1.1cm} | C{1.75cm} | C{2.25cm} | C{1.95cm} | C{2.39cm} | C{2.15cm}}
    \hline
    Model & \emph{SingleImage} & \emph{SingleImage-IV} & \emph{Stereo-FA-IV} & \emph{Stereo-FA-dp-IV} & \emph{Stereo-FA-MV} \\ \hline 
    \emph{MSL} & 426 & 424 & 410 & \textbf{407} & 417 \\ 
    \emph{MVL} &2033 &1121 &1028 &1022 & \textbf{1014}\\ 
    \emph{MCL} & \textbf{424153} & 485089 & 481056 & 478413 & 445336\\ 
    \hline
    \end{tabular}
    }
\label{tab:eval1}
\end{table}

\begin{figure}
\begin{center}
   \includegraphics[width=0.8\linewidth]{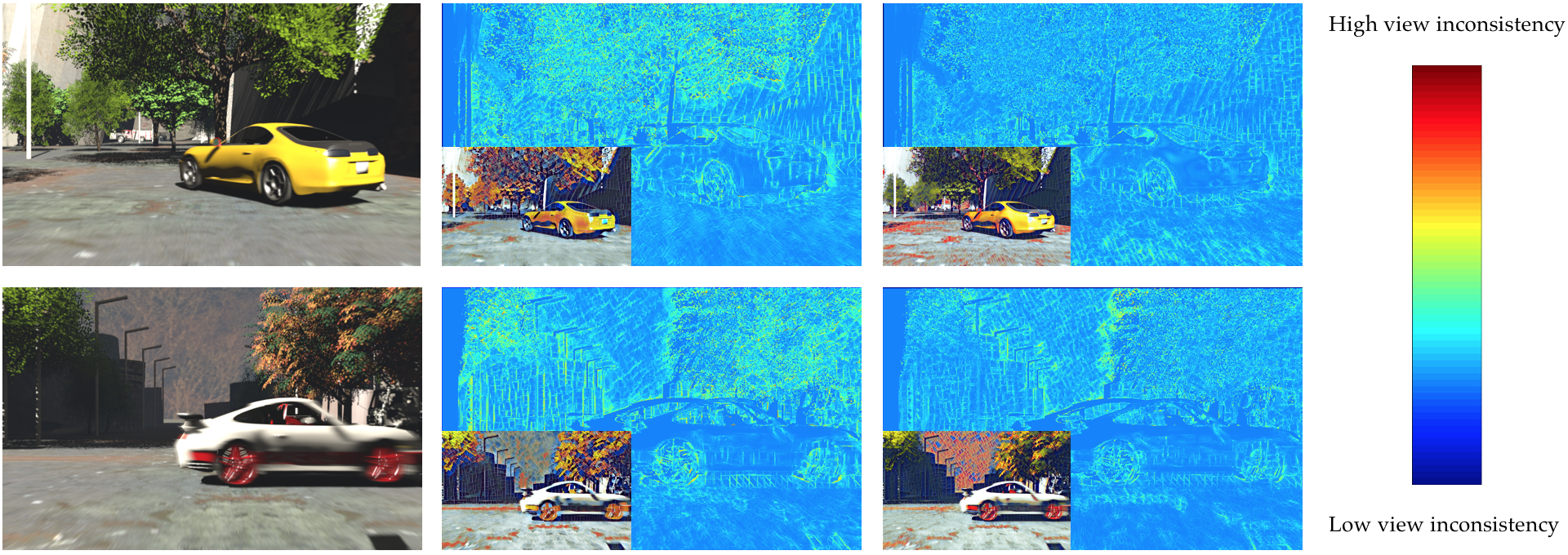}
\end{center}
   \caption{Visualization of the view inconsistency. The second column shows view inconsistency maps of the single-image style transfer method~\cite{johnson2016perceptual}. The third column shows our results. The last column is the color map of view inconsistency maps. Obviously, our results are more view-consistent.}
\label{fig:v}
\end{figure}

In order to give a more intuitive comparison, we visualize the view inconsistency maps  of the single image style transfer method and our proposed method in Fig. \ref{fig:v}. The view inconsistency map is defined as:
\begin{equation}
V^L = \sum\limits_{c}\left|\widehat{x}^{L}_c - W(\widehat{x}^{R})_c\right|\odot M^L,
\end{equation}
where $\widehat{x}^{L}_c$ and $W(\widehat{x}^{R})_c$ denote $c$-th channel of $\widehat{x}^{L}$ and $W(\widehat{x}^{R})$ respectively. $M$ is the per-pixel confidence mask of disparity map which is illustrated in Sec.\ref{sec:VL} . Note that $W$ denotes the warp operation using the ground-truth disparity map, provided by the Scene Flow Datasets~\cite{MIFDB16}. Compared with the results of \emph{SingleImage}, a larger number of blue pixels in our results indicate that our method can preserve the view consistency better. 

Moreover, a user study is conducted to compare \emph{SingleImage} with our method. Specifically, a total number of 21 participants take part in our experiment. Ten stereo pairs are randomly picked up from the Waterloo-IVC 3D database~\cite{wang2015quality}. For each of the stereo pair, we apply style transfer using three different style images (\emph{candy, fish, mosaic}). As a result, $3 \times 10$ stylized stereoscopic pairs are generated for each model. Each time, a participant is shown the stylized results of the two methods on a 3D TV with a pair of 3D glasses, and asked to vote for the preferred one (which is more view-comfortable). Specifically, the original stereo pairs are shown before the stylized results of the two methods, in order to give participants the correct sense of depth as references. Tab. \ref{tab:user} shows the final results. 73\% votes are cast to the stylized results generated by our method, which demonstrates that our method achieves better view consistency and provides more satisfactory visual experience.

\begin{table}
\caption{User preferences.}
\small
\begin{center}
\begin{tabular}{ c||c|c|c }
\hline
Style & Prefer ours & Prefer Johnson \etal's & Equal\\ \hline
\emph{Candy} & \textbf{143} & 29 & 38 \\
\emph{Fish} & \textbf{166} & 14 & 30 \\
\emph{Mosaic} & \textbf{152} & 24 & 34 \\
\hline
\end{tabular}
\end{center}
\label{tab:user}
\end{table}

\subsection{Ablation Study on Feature Aggregation}
To testify the effectiveness of the proposed feature aggregation block, we set up an ablation study. Our feature aggregation block consists of three key operations: warping, gating and concatenation. We test 3 variant models with different settings of these key operations for obtaining the final aggregated feature maps $\mathcal{A}^L$ and $\mathcal{A}^R$. For simplicity, we only describe the process of obtaining $\mathcal{A}^L$. 

The first model is \emph{SingleImage-IV}, where the single image method trained with image-level view loss and perceptual loss.
In the second model \emph{CON-IV}, $\mathcal{A}^L$ is obtained by concatenating $\mathcal{F}^R$ with $\mathcal{F}^L$. The last model \emph{W-G-CON-IV} uses our proposed feature aggregation block, which is equal to \emph{Stereo-FA-IV} as mentioned before.
Here we consider warping-gating as an indivisible operation, as the warping operation will inevitably introduce hollow areas in the occluded region, and the gating operation is used to localize the hollow areas and guide a feature aggregation process to fill the holes. All models above are trained with the perceptual loss and view loss, using \emph{Fish}, \emph{Mosaic}, \emph{Candy} and \emph{Dream} as the reference style images.

Tab. \ref{tab:eval2} shows the mean view loss of the 3 variant models. Comparing \emph{CON-IV} with \emph{SingleImage-IV}, we can see that concatenating $\mathcal{F}^R$ with $\mathcal{F}^L$ does help the decrease of the \emph{MVL}, which demonstrates that the concatenated skip connection is essential.
Comparing \emph{W-G-CON-IV} with \emph{CON-IV}, \emph{W-G-CON-IV} achieves better performance. This is because that $\mathcal{F}_r^R$ is aligned with $\mathcal{F}^L$ along the channel axis, which relieves the need of learning pixel correspondences. 

\begin{table}
\caption{$MVL$, $MSL$ and $MCL$ of three different feature aggregation blocks. Our proposed feature aggregation block architecture achieves the smallest $MVL$ and $MCL$, indicating the best view consistency and content preservation.}
\centering
    \begin{tabular}{ C{1.1cm} | C{2.25cm} | C{1.3cm} | C{2.39cm}}
    \hline
    Model & \emph{SingleImage-IV} & \emph{CON-IV} & \emph{W-G-CON-IV}  \\ \hline 
    \emph{MSL} & 424 &\textbf{328} & 410 \\ 
    \emph{MVL} &1121 &1068 &\textbf{1028}\\ 
    \emph{MCL} & 485089 & 489555 & \textbf{481056}\\ 
    \hline
    \end{tabular}
\label{tab:eval2}
\end{table}

\begin{figure}[t]
\begin{center}
   \includegraphics[width=0.7\linewidth]{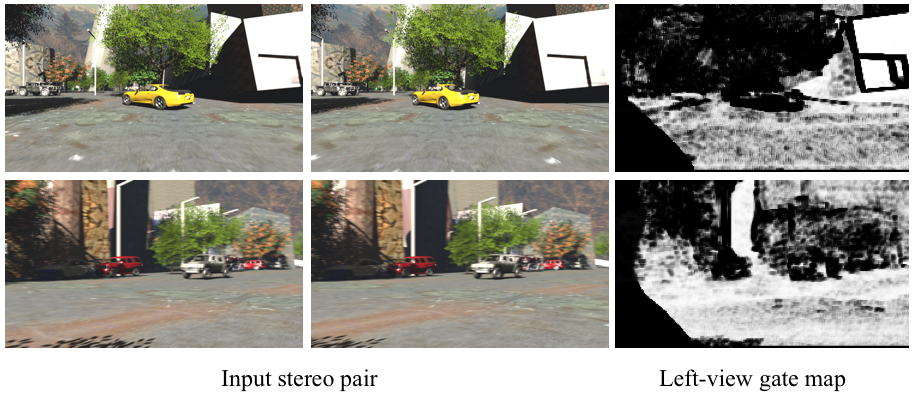}
\end{center}
   \caption{Visualization of gate maps. The left and middle columns are two input stereo pairs. The right column shows the left-view gate map generated by the gate sub-network.}
\label{fig:gate}
\end{figure}

In order to give an intuitive understanding of the gate maps, we visualize several gate maps in Fig.\ref{fig:gate}. Recalling that the Equation~\ref{eq:refine}, the refined feature map $\mathcal{F}_r^R$ is a linear combination of the initial feature map  $\mathcal{F}^L$ and the warped feature map  $W'(\mathcal{F}^R)$, under the guidance of the gate map. For simplicity, we only illustrate the gate maps for the left view. Generated gate maps are shown in  the right column.  The black regions in the gate maps indicate the mismatching between $\mathcal{F}^L$ and $W'(\mathcal{F}^R)$. Here, the mismatching is caused by occlusion and inaccurate disparity estimation. For the mismatched areas, the gate sub-network learns to predict 0 values to enforce  the refined feature map $\mathcal{F}_r^R$ directly copy  values from $\mathcal{F}^L$ to  avoid inaccurately incorporating information from the occluded regions in the right view. 

\section{Conclusion}
In this paper, we proposed a novel dual path network to deal with style transfer on stereoscopic images. While each view of an input stereo pair has been processed in an individual path to transfer the style from a reference image, a novel feature aggregation block was proposed to propagate the information from one path to another. Multiple feature aggregation strategies were investigated and compared to demonstrate the advantage of our proposed feature aggregation block.  To coordinate the learning of both the paths for gaining better view consistency, a multi-layer view loss was introduced to constrain the stylized outputs of both views to be consistent in multiple scales. The extensive experiments demonstrate that our method is able to yield stylized results with better view consistency than those achieved by the previous methods.

\clearpage

\bibliographystyle{splncs04}
\bibliography{egbib}

\begin{thebibliography}{10}
\providecommand{\url}[1]{\texttt{#1}}
\providecommand{\urlprefix}{URL }
\providecommand{\doi}[1]{https://doi.org/#1}

\bibitem{Basha2011Geometrically}
Basha, T., Moses, Y., Avidan, S.: Geometrically consistent stereo seam carving.
  In: Proceedings of ICCV (2011)

\bibitem{Butler:ECCV:2012}
Butler, D.J., Wulff, J., Stanley, G.B., Black, M.J.: A naturalistic open source
  movie for optical flow evaluation. In: {A. Fitzgibbon et al. (Eds.)} (ed.)
  European Conf. on Computer Vision (ECCV). pp. 611--625. Part IV, LNCS 7577,
  Springer-Verlag (Oct 2012)

\bibitem{chang2011content}
Chang, C.H., Liang, C.K., Chuang, Y.Y.: Content-aware display adaptation and
  interactive editing for stereoscopic images. IEEE Transactions on Multimedia
  \textbf{13}(4),  589--601 (2011)

\bibitem{chen2017coherent}
Chen, D., Liao, J., Yuan, L., Yu, N., Hua, G.: Coherent online video style
  transfer. In: Proceedings of ICCV (2017)

\bibitem{flickr}
Flickr: Flickr. \url{https://www.flickr.com}

\bibitem{gatys2016image}
Gatys, L.A., Ecker, A.S., Bethge, M.: Image style transfer using convolutional
  neural networks. In: Proceedings of CVPR (2016)

\bibitem{gupta2017characterizing}
Gupta, A., Johnson, J., Alahi, A., Fei-Fei, L.: Characterizing and improving
  stability in neural style transfer. In: Proceedings of ICCV (2017)

\bibitem{htcvive}
HTC: {HTC Vive}. \url{https://www.vive.com/us/}

\bibitem{huang2017real}
Huang, H., Wang, H., Luo, W., Ma, L., Jiang, W., Zhu, X., Li, Z., Liu, W.:
  Real-time neural style transfer for videos. In: Proceedings of CVPR (2017)

\bibitem{Huang_2017_ICCV}
Huang, X., Belongie, S.: Arbitrary style transfer in real-time with adaptive
  instance normalization. In: Proceedings of ICCV (2017)

\bibitem{imax}
IMAX: {IMAX}. \url{https://www.imax.com}

\bibitem{johnson2016perceptual}
Johnson, J., Alahi, A., Fei-Fei, L.: Perceptual losses for real-time style
  transfer and super-resolution. In: Proceedings of ECCV (2016)

\bibitem{kim2013stereoscopic}
Kim, Y., Lee, Y., Kang, H., Lee, S.: Stereoscopic 3d line drawing. ACM
  Transactions on Graphics (TOG)  \textbf{32}(4), ~57 (2013)

\bibitem{kingma2014adam}
Kingma, D., Ba, J.: Adam: A method for stochastic optimization. arXiv preprint
  arXiv:1412.6980  (2014)

\bibitem{kooi2004visual}
Kooi, F.L., Toet, A.: Visual comfort of binocular and 3d displays. Displays
  \textbf{25}(2),  99--108 (2004)

\bibitem{lee2012scene}
Lee, K.Y., Chung, C.D., Chuang, Y.Y.: Scene warping: Layer-based stereoscopic
  image resizing. In: Proceedings of CVPR (2012)

\bibitem{3dtv}
LG: {4K HDR Smart TV}. \url{http://www.lg.com/us/tvs/lg-OLED65G6P-oled-4k-tv}

\bibitem{li2017demystifying}
Li, Y., Wang, N., Liu, J., Hou, X.: Demystifying neural style transfer. arXiv
  preprint arXiv:1701.01036  (2017)

\bibitem{li2017diversified}
Li, Y., Fang, C., Yang, J., Wang, Z., Lu, X., Yang, M.H.: Diversified texture
  synthesis with feed-forward networks. arXiv preprint arXiv:1703.01664  (2017)

\bibitem{luo2015geometrically}
Luo, S.J., Sun, Y.T., Shen, I.C., Chen, B.Y., Chuang, Y.Y.: Geometrically
  consistent stereoscopic image editing using patch-based synthesis. IEEE
  transactions on visualization and computer graphics  \textbf{21}

\bibitem{holo}
Microsoft: {Microsoft HoloLens}. \url{https://www.microsoft.com/en-gb/hololens}

\bibitem{MIFDB16}
N.Mayer, E.Ilg, P.H{\"a}usser, P.Fischer, D.Cremers, A.Dosovitskiy, T.Brox: A
  large dataset to train convolutional networks for disparity, optical flow,
  and scene flow estimation. In: Proceedings of CVPR (2016)

\bibitem{northam2012consistent}
Northam, L., Asente, P., Kaplan, C.S.: Consistent stylization and painterly
  rendering of stereoscopic 3d images. In: Proceedings of NPAR (2012)

\bibitem{ruder2016artistic}
Ruder, M., Dosovitskiy, A., Brox, T.: Artistic style transfer for videos. In:
  Proceedings of GCPR (2016)

\bibitem{simonyan2014very}
Simonyan, K., Zisserman, A.: Very deep convolutional networks for large-scale
  image recognition. arXiv preprint arXiv:1409.1556  (2014)

\bibitem{stavrakis2005image}
Stavrakis, E., Bleyer, M., Markovic, D., Gelautz, M.: Image-based stereoscopic
  stylization. In: Image Processing, 2005. ICIP 2005. IEEE International
  Conference on. vol.~3, pp. III--5. IEEE (2005)

\bibitem{ulyanov2016texture}
Ulyanov, D., Lebedev, V., Vedaldi, A., Lempitsky, V.S.: Texture networks:
  Feed-forward synthesis of textures and stylized images. In: Proceedings of
  ICML (2016)

\bibitem{ulyanov2016instance}
Ulyanov, D., Vedaldi, A., Lempitsky, V.: Instance normalization: The missing
  ingredient for fast stylization. arXiv preprint arXiv:1607.08022  (2016)

\bibitem{wang2017zm}
Wang, H., Liang, X., Zhang, H., Yeung, D.Y., Xing, E.P.: Zm-net: Real-time
  zero-shot image manipulation network. arXiv preprint arXiv:1703.07255  (2017)

\bibitem{wang2015quality}
Wang, J., Rehman, A., Zeng, K., Wang, S., Wang, Z.: Quality prediction of
  asymmetrically distorted stereoscopic 3d images. IEEE Transactions on Image
  Processing  \textbf{24}(11),  3400--3414 (2015)

\bibitem{Wang_2017_CVPR}
Wang, X., Oxholm, G., Zhang, D., Wang, Y.F.: Multimodal transfer: A
  hierarchical deep convolutional neural network for fast artistic style
  transfer. In: Proceedings of CVPR (2017)

\end{thebibliography}

\end{document}